# Automated Solubility Analysis System and Method Using Computer Vision and Machine Learning

Gahee Kim,[a] Minwoo Jeon,[b] Hyun Do Choi,[a] Jun Ki Cho,[a] Youn-Suk Choi,[a] and Hyoseok Hwang*,[b]

[a] Materials Research Center, Samsung Advanced Institute of Technology, Samsung Electronics, Suwon-si, Republic of Korea

[b] Department of Software Convergence, Kyung Hee University, Yongin-si, Republic of Korea
E-mail: hyoseok@khu.ac.kr
Phone: +82 (0)31 2013749

**Abstract:** In this study, a novel active solubility sensing device using computer vision is proposed to improve separation purification performance and prevent malfunctions of separation equipment such as preparative liquid chromatographers and evaporators. The proposed device actively measures the solubility by transmitting a solution using a background image. The proposed system is a combination of a device that uses a background image and a method for estimating the dissolution and particle presence by changing the background image. The proposed device consists of four parts: camera, display, adjustment, and server units. The camera unit is made up of a rear image sensor on a mobile phone. The display unit is comprised of a tablet screen. The adjustment unit is composed of rotating and height-adjustment jigs. Finally, the server unit consists of a socket server for communication between the units and a PC, including an automated solubility analysis system implemented in Python. The dissolution status of the solution was divided into four categories and a case study was conducted. The algorithms were trained based on these results. Six organic materials and four organic solvents were combined with 202 tests to train the developed algorithm. As a result, the evaluation rate for the dissolution state exhibited an accuracy of 95 %. It was confirmed that the parameters used in the algorithm moved in a specific direction even during the real-time measurement. This indicates that our devices and methods work well for solubility sensing. In addition, the device and method must develop a feedback function that can add a solvent or solute after dissolution detection using solubility results for use in autonomous systems, such as a synthetic automation system. To implement this function, improving the detection limit and measurement rate using various hardware functions is necessary. In addition, improvements to elaborate algorithms using numerous data should be made. Finally, the diversification of the sensing method is expected to extend not only to the solution but also to the solubility and homogeneity analysis of the film.

# 1  Introduction

Solubility is defined as the maximum amount of a substance dissolved in a given amount of solvent under the given process conditions when the system is in equilibrium. All chemical reactions begin with the reactants uniformly dissolved in the solvent; therefore, the exact measurement of solubility is essential [1-3]. When the reactant is uniformly dissolved in the solvent, its intrinsic functionality can be completely maintained, and the product can be obtained through the preferred mechanism. Solubility measurements have significant importance, not only in the initial stages of chemical reactions but also in the intermediate stages. The solubility of the intermediate product can be helpful in determining whether to continue the reaction [4-6]. Furthermore, solubility data can provide crucial information about the appropriate solvent and the amount of solvent for recrystallization during the work-up and purification stages.

Solubility can be measured through optical density measurements using a spectrometer,

turbidity measurements using a nephelometer and turbidimeter, and high-performance liquid chromatography (HPLC) [7-15]. Optical density spectrometry and turbidity measurements using a nephelometer and a turbidimeter are commonly used methods for solubility measurements. A spectrometer measures the concentration of a solution by measuring the transmittance and reflectance of the solution using light. A nephelometer and a turbidimeter in a measure the turbidity of the solution based on the transmission and scattering of light. The turbidity of a liquid is determined by the presence of finely dispersed suspended particles. If a beam of light passes through a turbid sample, its intensity is reduced due to scattering, and the quantity of scattered light depends on the concentration and size distribution of the particles. All these methods use light scattering and require separate sampling and preparation processes; therefore, solubility analysis through real-time measurement is not possible. Most automated solubility analysis systems are based on HPLC [9,16-18], which is a reliable and standard method for quantification; however, it has some disadvantages, such as the requirement of a reference material for evaluating the materials, calibration for automation, and a significant amount of time for accurate analysis. The most significant disadvantage of solubility measurement via the three methods mentioned above is the use of invasive methods, such as the collection of samples for solubility measurement or direct injection of the analyzer into the sample. Real-time analysis using invasive methods is impossible, making them suitable only for automated or autonomous systems. This study proposes a new analytical device and method using computer vision to sense solubility in real-time.

Recently, studies on solubility analysis using computer vision have been reported [1,19,20]. The vision method does not require sampling; therefore, solubility can be measured non-invasively, and stopping the reaction or disrupting the environment in the reaction state for the solubility analysis is not required. In other words, real-time analysis is possible using computer vision. In addition, we developed our own image recognition method and devised a method for measuring solubility using a wide region of interest (ROI) for the first time. In other words, although the existing solubility analysis methods provide solubility information for a microscopic area, our method can provide that for the entire sample. Our solubility-sensing module consisted of camera, display, adjustment, and server units. The image sensor in the camera unit was used to capture an image of the flask, onto which the background image of the display unit was projected. The captured image was automatically transmitted to the personal computer (PC), the server unit, via the socket server. The captured images were preprocessed, and features were extracted using a computer vision algorithm. Finally, using the extracted features, a support vector machine (SVM) [21-23] classifier was trained, and the solution was classified. Our new method cannot provide an exact numerical value for solubility; however, with a picture of the solution, it is possible to accurately determine whether the solute is dissolved in real-time.

## 2  Materials and Methods

## 2.1 Proposed system and method

The proposed device includes a camera, display, adjustment, and server units, as shown in Fig. 1a. The camera unit consists of a rear image sensor on a mobile phone and captures an image, which encompasses the characteristics of the solution, as shown in Fig. 1a. The display unit consists of a tablet screen, which displays a check and white background image. The image then passes through the flask toward the detection camera. The third component is the adjustment unit, which is composed of rotating and height-adjustment jigs. The rotating jig rotates and clamps the flask to capture another image of the solution such that undissolved particles can be detected. The height-adjustment jig was used to adjust the height of the phone or tablet to capture photographs of the flask from different heights. Finally, the server unit automatically feeds the results by determining the degree of dissolution through automated solubility analysis system (ASAS) implemented in Python. The images captured by the camera unit were sent to the server and classified after performing ASAS calculations.

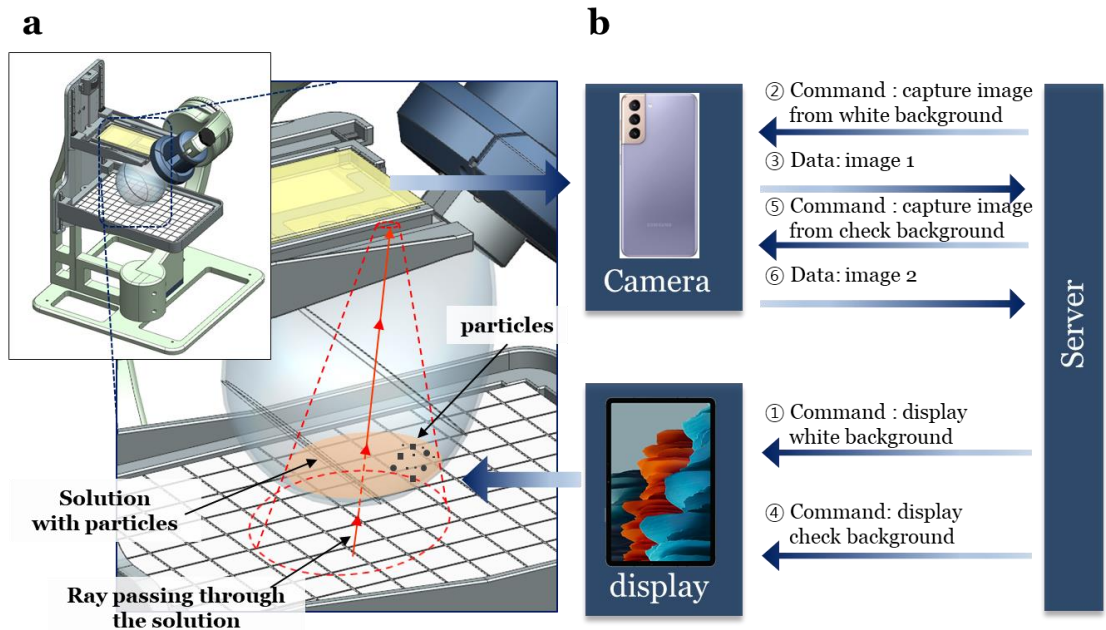

## 2.2 Measurement process

In our proposed measurement process, a communication process was established between the camera, display, and server units, as illustrated in Fig. 1b. Communication among the camera, display, and server units was performed through the socket server. First, the server commands the display unit to display a white background image and then commands the camera unit to capture the flask with the white background image. The captured image is sent back to the server to acquire features through ASAS. Similarly, the server commands the display unit to display a check background image and then commands the camera unit to capture the flask with the check background image. The captured image is then sent back to the server to acquire features through ASAS. The acquired features are used to train the SVM classifier and measure the solubility. This process is performed sequentially to automatically measure the solubility of the solution.

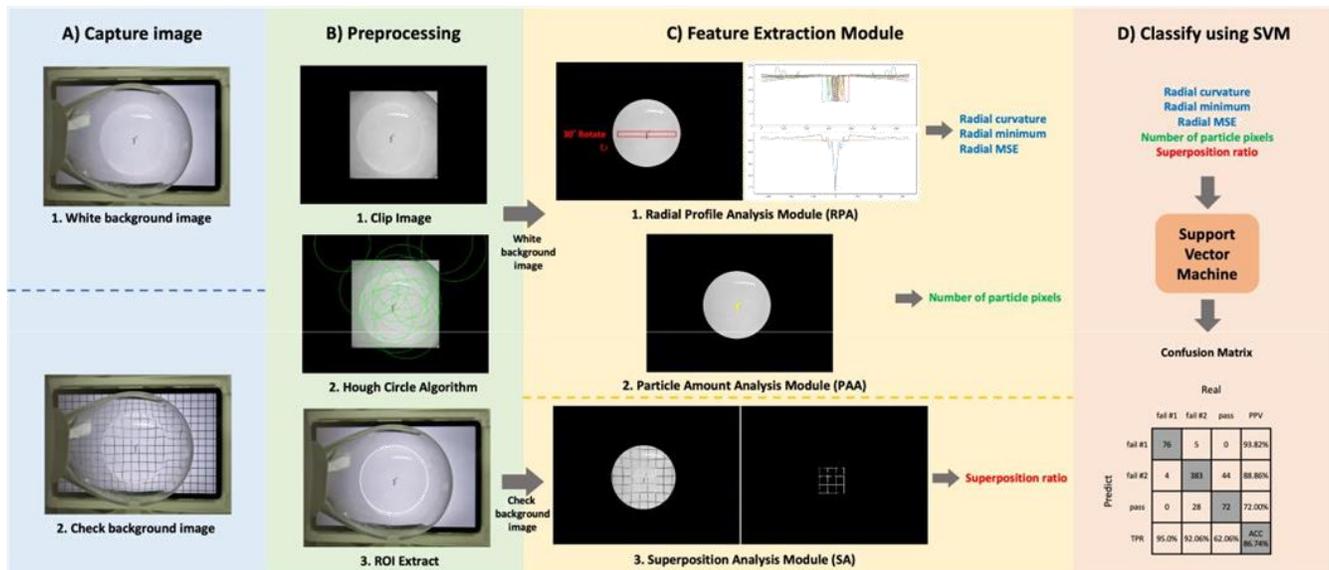

## 2.3 Analysis method

For automated solution analysis, the image to be analyzed must be captured. The flask was fixed to a tablet that displayed a background image and captured the solution in the flask, onto which the background image was projected. There were two types of background images: white and check patterns. The solubilities of the solutions were classified according to the workflow shown in Fig. 2. In the preprocessing method, an ROI was identified in the original image, which corresponded to the solution in the image. The analysis of only the ROI can reduce the computation time and enhance the performance of the system. First, 900 pixels were clipped horizontally and vertically from the center of the original image. Because the solution was captured at the center of the image, only the center of the image was required to obtain the ROI. As the solution was captured as a circle, the ROI was obtained using the Hough circle [24] algorithm. In the feature extraction method, the ROI was received as an input and five features were extracted using the radial profile analysis (RPA), particle amount analysis (PAA), and superposition analysis (SA) modules. The five features are listed in Table 1. RPA and PAA use the white background image to detect undissolved solute particles. The SA module uses the check background image to obtain the turbidity of the solution based on the degree to which the check is covered. The RPA rotates the image by 30 ° based on the center coordinates of the image, obtains pixel intensity distribution corresponding to the diameter of a circle in a radial form every time it rotates, and obtains a total of 12 distributions. The average of the 12 radial distributions is calculated and the average is approximated as a quadratic function ($ax^2 + c = 0$). The parameters 'a' and 'c' in a quadratic function represent the curvature and the minimum value, respectively, and are used as features. In addition, the mean squared error (MSE) between the approximated quadratic function and the average of the radial distributions is also used as a feature. The PAA module obtains the number of pixels evaluated undissolved solutes through adaptive thresholding [25] and uses it as a feature. The SA module extracts lines through Canny edge detection [26] and a progressive probabilistic Hough transform [27,28] on the image to which adaptive thresholding is applied. Furthermore, it extracts the check patterns through morphological expansion and closure operations [29]. The superposition of the extracted check pattern and the ground truth check pattern was used as a feature. After training the SVM with the five features and corresponding pre-labeled ground truth for each image, the results are categorized into three groups: Pass, Fail 1, and Fail 2 (Fig. 2d).

Table 1. Five features obtained through the automated solubility analysis system.

| Feature. | Background type | Feature extraction Module | Objective |
|---|---|---|---|
| Radial curvature | White | RPA | Particle detection |
| Radial minimum | White | RPA | Particle detection |
| Radial MSE | White | RPA | Particle detection |
| Number of particle pixels | White | PAA | Particle detection |
| Superposition ratio | Check | SA | Turbidity measurement |

## 2.4 Preparation of the solubility test

Generally, the solubility of a substance is defined as the number of grams of the substance that dissolves in 100 g of water at a given temperature [1,2]. To show how the solubility varies with temperature, solubility curves were plotted with temperature on the x-axis and solubility (in g per 100 g of solvent) on the y-axis. The line on the solubility curve represents solution saturation. Below these lines, the solution was not saturated, and more solute could be dissolved. Above these lines, the solution contains more solute than can be dissolved in the solvent at the given temperature; therefore, some of it remains as undissolved particles or crystals. The solubility was lower at lower temperatures in the solubility curve. The solute molecules combined with the solvent molecules were again combined with other solute molecules at a lower temperature until the solution became cloudy or precipitation began to occur. A new curve, termed as the supersolubility curve, was plotted above the solubility curve, and the area between the solubility and supersolubility curves was defined as the metastable zone width (MSZW) [30-33]. The solution in the MSZW is in a metastable state, and slight external shocks, such as temperature and pressure changes, pH changes, and dust, can change the state of the solution to a clear or precipitation state. Based on the basic theory of the solubility curve, an algorithm that distinguishes the three states of clearness (Pass), turbidity (Fail 1), and particles (Fail 2) in a solution was applied to determine whether the solution was dissolved.

## 2.5 Device setup

Our proposed device consists of camera, display, and adjustment units, as shown in Fig. 3. The camera unit used the rear 12 MP wide-angle camera of Samsung Galaxy S21 and captured images at 1920 × 1440 pixels to reduce the amount of data computation. We installed a self-made android application to communicate smoothly with the server and automatically shoot and adjust the focus. The display unit consisted of a Samsung Galaxy tab S7 FE with a display resolution of 2560 × 1600 pixels. We installed our in-house developed android application to communicate seamlessly with the server and automatically display the desired background image. The adjustment unit consisted of a rotating jig and a height-adjustment jig fabricated using a 3D printer. The rotating jig can be rotated by 360 °, and the height-adjustment jig can be adjusted to a height of up to several centimeters. The server unit consisted of a PC and a socket server. The socket server is designed to communicate between the PC and camera unit or the PC and display unit.

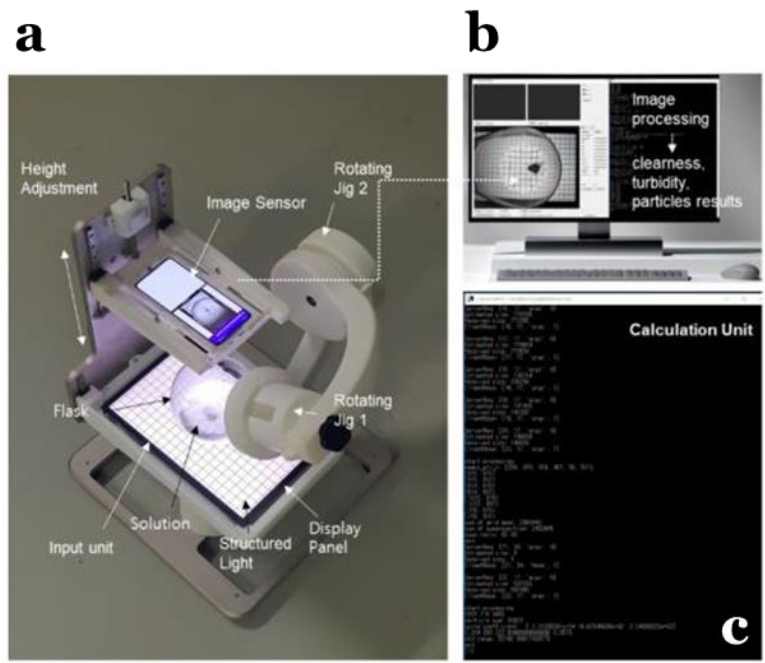

**2.6   Case study for machine learning**

To create an algorithm that determines the solubility status, a case study was performed manually for four typical cases, which are as follows. (A) The solutes with very high solubility appeared clear in the solvent. (B) The solutes dissolved well in the solvent until the solution reached saturation at a certain concentration where it appeared as undissolved particles in the solvent. (C) The solutes remained undissolved in the solvent when sunk to the bottom of the flask and attached to the inner wall of the flask, or when water droplets were formed in the inner wall of the flask as a residue. (D) The solutes with very low solubility in the solvent appeared as floating materials; consequently, the turbidity of the solution was very high. Deionized (DI) water was used as the solvent, and copper sulfide, copper acetic acid, copper bromide, and palladium acetic acid were used to prepare samples for the cases A, B, C, and D, respectively. Images of the DI water and the four different solutes in the flask are shown in Fig. 4. Each sample was prepared at different concentrations; that is, the concentrations of $CuSO_4$ were 5 g/100 mL, 10g/100 mL, and 15 g/100 mL; of CuOAc were 0.5g/100mL, 4g/100mL, 6g/100mL, and 7.03 g/100mL; of CuBr were 0.01g/100mL, 0.1g/100mL, 0.5g/100mL, and 1g/100mL; and of PdOAc were 0.01g/100mL, 0.1g/100mL, 0.5g/100mL, and 1g/100mL. For each test sample, two images were obtained using background displays of the check and white patterns, and the obtained data were analyzed to learn the data. The process of analyzing the image is as follows. First, the display panel and camera were calibrated. The first background pattern was then displayed on the display panel, and the image was captured and saved. Next, the second background pattern was displayed on the display panel, and the image was captured and saved. Finally, both images were analyzed simultaneously. The images obtained using each pattern were analyzed and learned using different algorithms after the ROI extraction.

## 2.7 Solubility analysis using the developed algorithm

The image analysis algorithm was developed through experiments using the four cases mentioned above, and 194 cases from 97 tests for each pattern were analyzed with a combination of five organic substances and three solvents to verify the developed algorithm. The five organic substances used were 2-bromo-4-phenylpyridine, 4-methoxyphenol, naphthalic anhydride, 2,2-bithiophene5carboxaldehyde, and 4,4-bis(a,adimethylbenzyl)diphenylamine. Furthermore, the three solvents were toluene, methylene chloride, and hexane. The data for each combination and the molar concentrations are listed in Table 2. The sample of such combinations was analyzed using our solubility sensing device independently to transmit the background image to determine whether it was dissolved using the developed algorithm. The used background images had white and check pattern background, and the images were evaluated as Pass (clear) and Fail 1(turbidity) /Fail 2 (particles in solution) to create an algorithm that determines the soluble status. Further, the case study was progressed manually.

## 3 Results and discussion

### 3.1 Case study for machine learning

The SVM classifier was trained using the ground truth and five features obtained by applying the feature-extraction module to the solution image. We constructed a training dataset consisting of solution images and ground truths to train the SVM classifier. The ground truth was manually determined as Fail 1, Fail 2, and Pass according to the dissolution state of the solution. For example, when the solute was completely dissolved and transparent, it was categorized as Pass. When the solute remained in the form of particles without dissolution, it was categorized as Fail 2. When the solute did not dissolve and the solution became cloudy, the solute was categorized as Fail 1. Fail 1 and Fail 2 are collectively referred to as Fail. The images in the training dataset were composed of solutions containing $CuSO_4$, CuOAc, CuBr, and PdOAc, and a total of 153 cases were captured. The 153 cases consisted of 20 cases of Fail 1, 104 cases of Fail 2, and 29 cases of Pass. Each case was captured against white and checked backgrounds. To improve the generalization performance of the SVM, the size of the dataset was increased through horizontal flip, vertical flip, and horizontal and vertical flip image augmentation. Consequently, a dataset with 612 cases and 1224 images was constructed. We constructed the dataset using images directly captured in the laboratory, so the dataset is limited in quantity. Therefore, we did not create an additional test dataset. The results of classifying the dataset solution images in the trained SVM were expressed using a confusion matrix. In the case of Fail 1, 4 of the 80 images were misclassified. In the case of Fail 2, 33 of 416 images were misclassified. In the case of Pass, 44 of 116 images were misclassified. The accuracies of Fail 1, Fail 2, and Pass were 95 %, 92.06 %, and 62.06 %, respectively. In addition, a four-fold validation process was performed to prove the generalization performance of the SVM. The four-fold validation results

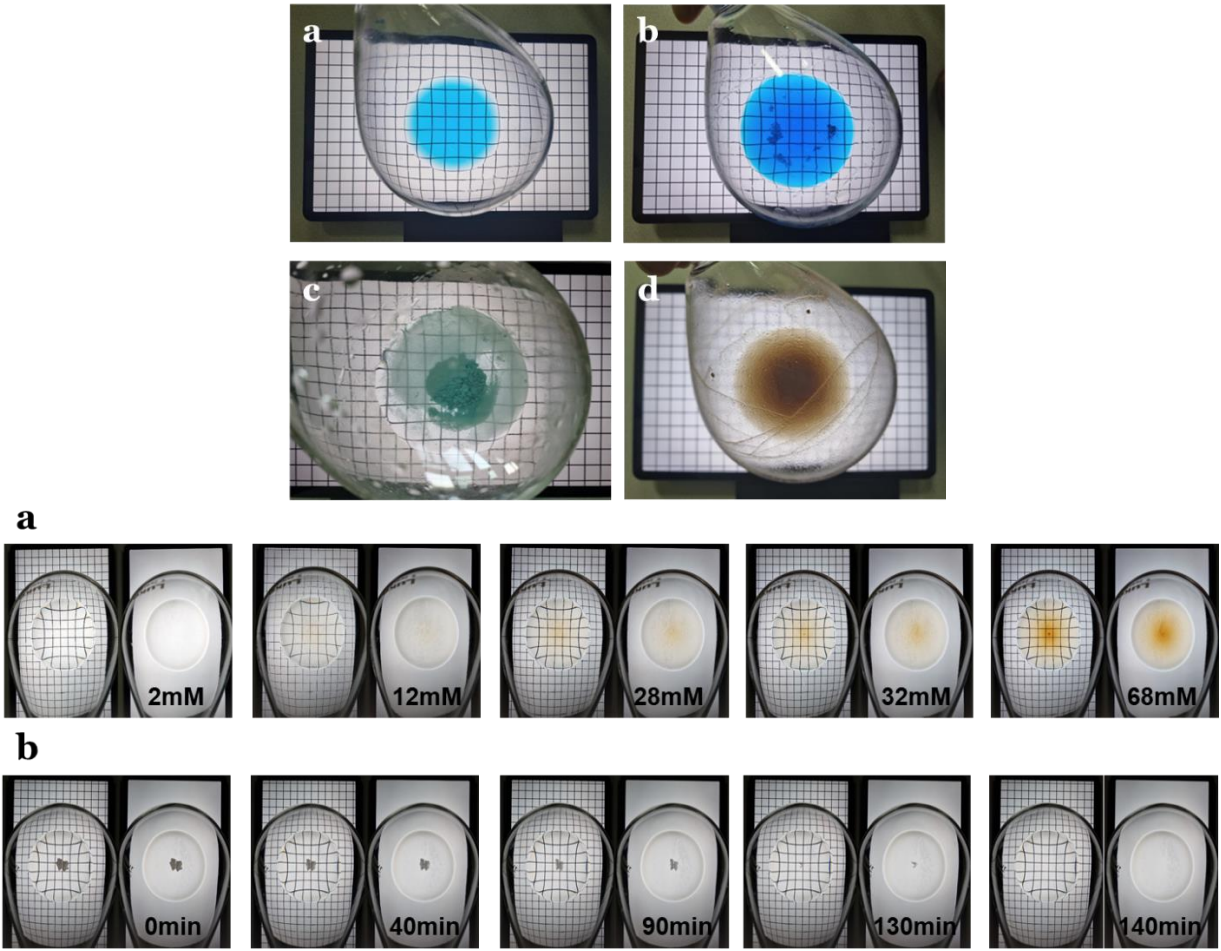

are listed in Table 3.

Table 2. Combination of solute and solvent

| Test No. | Solute | Solvent | Hit No. | Error No. | Accuracy (%) |
|---|---|---|---|---|---|
| 1 | 2-bromo-4-phenylpyridine | Toluene | 42 | 2 | 95.2 |
| 2 | 4-Methoxyphenol | Toluene | 6 | 0 | 100.0 |
| 3 | | MC | 6 | 0 | 100.0 |
| 4 | Naphthalic anhydride | Hexane | 6 | 0 | 100.0 |
| 5 | | Toluene | 6 | 3 | 50.0 |
| 6 | 2,2'-Bithiophene-5-carboxaldehyde | Toluene | 20 | 2 | 90.0 |
| 7 | 4,4'-bis(a,a-dimethylbenzyl)diphenylamine | Toluene | 27 | 2 | 92.6 |

Table 3. Four-fold validation results.

| Train Dataset | Accuracy (%) |
|---|---|
| Fold 1 | 86.92 |
| Fold 2 | 81.04 |
| Fold 3 | 81.04 |
| Fold 4 | 87.58 |
| Average | 84.15 |

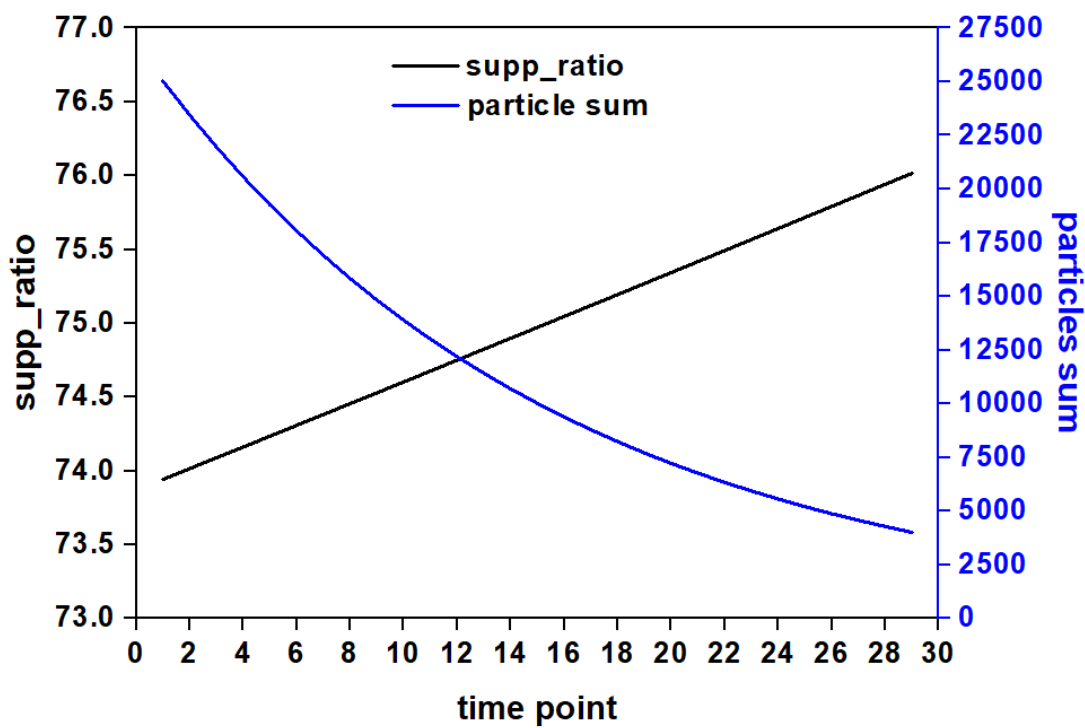

## 3.2 Solubility analysis using the developed algorithm

First, we briefly summarize the results obtained for the five organic materials. Here, we report the solubility analysis results of 2-bromo-4-phenylpyridine in toluene and 50 mM 4-methoxyphenol in toluene, and the results for the remaining materials are briefly provided in the Supplementary Information.

**3.2.1 2-bromo-4-phenylpyridine in toluene.** 2-bromo-4-phenylpyridine was increased to toluene at the same intervals, and the concentration was increased sequentially from 2 to 68 mM to analyze its dissolution in toluene. The number of analytical data points was 40, measured under 20 molar concentrations under two background images, which are shown in Fig 5a. At that time, the physical effects, such as heating, stirring, and shaking, were excluded. After 2-bromo-4-phenylpyridine was added to toluene, a background image was displayed to analyze the dissolution state in the order of white and checkered patterns. 2-bromo-4-phenylpyridine changed from Pass to Fail when its concentration in toluene was increased from 28 mM to 32 mM, and the results depended on the background pattern. The result of the analysis using the white background changed from Pass to Fail under lower molar concentration conditions compared with the result using the checkered pattern background. This result is attributed to the border present in the checkered pattern background, which helps to accurately distinguish the presence of particles on the grid or when the clearness of the grid is ambiguous. Next, the 194 cases tested using the developed algorithm were classified into four categories: A (52), B (100), C (26), and D (16). All of the dissolution statuses classified as A had to be assessed as Pass; however, 3 out of 52 were assessed as Fail, indicating 94.23 % accuracy of the developed algorithm. The dissolution state classified as B was expected to be Fail, and only one of the 100 cases was evaluated as Pass. In the case of C, all of them were expected to be Fail, and as expected, all 26 tests were assessed as Fail. This indicates that the accuracies of the algorithm in the cases of B and C were 99.00 % and 100.0 %, respectively. In the case of D, 1 out of 16 was assessed as Pass, and the accuracy was 93.75 %. Five out of the 194 tests were incorrectly judged, resulting in 97.42 % accuracy of the entire test. As mentioned previously, the developed algorithm can be divided into Fail 2, when the solute remains undissolved as particles, and Fail 1, which determines turbidity. No misjudgments were observed between Fail 1 and Fail 2, and all of them occurred between Pass and Fail 2; that is, three evaluated the Pass samples as Fail 2, and two tests evaluated the Fail 2 samples as Pass. The accuracies of these tests were calculated as 94.23 % and 98.28 %.

**3.2.2 4-methoxyphenol of 50 mM in toluene.** The 4-methoxyphenol (50 mM) was placed in a distilled flask with toluene and analyzed every 5 min until it was completely dissolved. The images are shown in Fig 5b. There were no changes to the environment, such as heating, stirring, or shaking. 4-Methoxyphenol was completely dissolved at room temperature for approximately 140 min, and it was confirmed that the analysis of the dissolved state using the developed solubility device was also evaluated as Pass from Fail at 140 min. Using these results, the analysis algorithm parameters of the two structured pattern-white backgrounds and the check pattern background were plotted as the time of solubility state analysis. The superposition ratio, which is a parameter used in the check pattern analysis, was calculated using the matching ratio of the sum of the grid mask and superposition. The larger the superposition ratio, the clearer the grid of the check pattern; therefore, it can be assumed that this value moves in an increasing direction when the solute is well dissolved in the solvent. The sum of the particles, which is a representative parameter of white background analysis, can be easily inferred to move in the direction of the smaller sum of the particles as they are dissolved. The results based on the times of these two parameters are shown in Fig. 6. The two results showed accurate results with time, as we assumed, and confirm the excellent performance of our developed algorithm in real-time. The plots for parameter 'c' and the parameters that assessed the homogeneity of the white background are shown in Supplementary Information.

## 4  Conclusions

In this study, we proposed a solubility analysis system to detect the dissolution state of a target solute. The system includes camera, display, adjustment, and server units. The display unit shows a white and checkered background image, and the image sensor of the camera unit captures the image passing through the flask with the target solution on the background image. The presence of particles and turbidity were calculated, and the solution was classified by solubility using a pretrained SVM classifier.

The captured image had features and characteristics different from those of the original image generated by the background display device, depending on the turbidity and particle size of the solution. The differences in the images were analyzed using the proposed algorithm to determine the solubility and presence or absence of particles. In 202 independent tests combining six organic substances and four organic solvents, a 95 % dissolved state accuracy was achieved.

We expect that the proposed system can be used in autonomous material development [1,34,35] for real-time solubility analysis, which determines the amount of additional supply of solvent or progress to the next process. In addition, the system can be used in most chemical processes in which chemists check the dissolution state with the naked eye, such as recrystallization, filtration, and solvent exchange. In future work, we will focus on applying the proposed system to these areas.

# 5   Table of Contents artwork

**Supporting Information**

Please find supplementary material provided with this file.

**Notes**
The authors declare no conflict of interest.

**Figure Captions**

Fig. 1: (a) Schematic illustration of the proposed solubility prediction system and (b) its operation block diagram.

Fig. 2: Overview of the automated solubility analysis system. The system is divided into four parts, including image capturing, preprocessing, feature extraction, and feature classification using SVM. The squares with different colors represent modules with different functions in the network. After the captured images undergo preprocessing, features are extracted using RPA and PAA for the white background image and SA for the check background image.

Fig. 3: (a) Photograph of the solubility sensing device. (b) PC system to control the hardware and software. (c) Screenshot of the image processing terminal.

Fig. 4: Four categories of dissolution for algorithm learning. (a) Category A: The solute is very well dissolved in the solvent, (b) category B: the solute is no longer dissolved beyond saturation, (c) category C: the solute is not dissolved at all in the solvent, and (d) category D: the solute is only partly dissolved and the solvent is very turbid.

Fig. 5: Changes in solubility (a) with increasing molar concentration and (b) at the same molar concentration.

Fig. 6: Changes in the superposition ratio and estimated number of the particles with respect to time. The superposition ratio between the background pattern and light passing through the solution linearly increases as 4-methoxyphenol dissolves in the solvent. In contrast, the estimated particle sum decreases gradually.